\documentclass[journal]{IEEEtran}
\ifCLASSINFOpdf
  \usepackage[pdftex]{graphicx}
  \usepackage[hidelinks]{hyperref}
  \usepackage{subfigure}
  \usepackage{hyperref}
\else
\fi
\usepackage{amsmath}
\usepackage{amssymb}
\usepackage{balance}
\begin{document}
%
\title{Extraction of Key-frames of Endoscopic Videos by using Depth Information}
%
%
%

\author{Pradipta Sasmal,\IEEEmembership{}
Avinash~Paul,\IEEEmembership{}
M.K.~Bhuyan,\IEEEmembership{~Senior Member,~IEEE}, and Yuji Iwahori
\thanks{P. Sasmal, Avinash Paul
       and~M.K.~Bhuyan are with the Department
of Electronics and Electrical Engineering, Indian Institute of Technology Guwahati, India.Yuji Iwahori is with Department of Computer Science, Chubu University, Kasugai, Japan.
*Author 1 and Author 2 have equal contribution.
E-mails: (s.pradipta, paul18, and mkb)@iitg.ac.in, iwahori@isc.chubu.ac.jp.}
}

%
%

\markboth{IEEE Signal Processing Letters}%
{Sasmal  \\MakeLowercase{\textit{et al.}}: Extraction of Key-frames of Endoscopic Videos by using Depth Information}
%



\maketitle

\begin{abstract}
A deep learning-based monocular depth estimation (MDE) technique is proposed for selection of most informative frames (key frames) of an endoscopic video. In most of the cases, ground truth depth maps of polyps are not readily available and that is why the transfer learning approach is adopted in our method. An endoscopic modalities generally capture thousands of frames. In this scenario, it is quite important to discard low-quality and clinically irrelevant frames of an endoscopic video while the most informative frames should be retained for clinical diagnosis. In this view, a key-frame selection strategy is proposed by utilizing the depth information of polyps.
%
%
In our method, image moment, edge magnitude, and key-points are considered for adaptively selecting the key frames. One important application of our proposed method could be the 3D reconstruction of polyps with the help of extracted key frames. 
Also, polyps are localized with the help of extracted depth maps.
\end{abstract}


%
\IEEEpeerreviewmaketitle

\section{Introduction}

Wireless Capsule Endoscopy (WCE) is a non-invasive modality to monitor the conditions of the internal viscera of a human body. WCE moves along the gastro-intestinal (GI) tract to capture images. It is extensively used to detect polyps in colon regions, which become cancerous if left untreated. Colorectal cancer is the third most prevalent cancer today \cite{siegel2017colorectal}. The capsule moves under the peristalsis movement, and it is very difficult to control the motion and orientation of the camera. Thus, redundant and clinically non-significant frames are generally obtained in a video sequence. 
WCE takes nearly 8 hours, capturing close to 50000 frames.
A large part of the data is clinically not significant and needs to be removed \cite{lee2013reducing}.

Several methods have been proposed for detection and localization of polyps in endoscopy frame \cite{li2012comparison}\cite{tjoa2002automated}.
A recent work focusing on video summarization instead of anomalies detection like bleeding or ulceration is proposed by Li \textit{et al.} \cite{li2010wireless}. Iakovidis \textit{et al.} \cite{iakovidis2010reduction} used clustering-based methods for video summarization. Researchers are working on visual attention models, like saliency maps for finding key-frames of videos \cite{hua2005generic}. 
Malignant polyps usually have a convex shape and are more textured compared to benign polyps. Getting a 3D view of the polyp surface can greatly help in resection \cite{law2016endoscopic}. A good 3D reconstruction of an object in an image entails dense depth estimation. The 3D view gives shape and size information of a polyp. Depth estimation of endoscopic images is a challenging task as the endoscopic images are monocular. 

Eigen \textit{et al.}, \cite{eigen2014depth} introduced a  multi-scale information approach which takes care of both global scene structure and local neighboring pixel information. A scale-invariant loss is used for MDE. Similarly, Xu \textit{et al.} \cite{ricci2018monocular} formulated MDE as a continuous random field problem (CRF). They fused the multi-scale estimation computed from the inner semantic layers of a CNN with a CRF framework. Instead of finding continuous depth maps, Fu et al.  \cite{fu2018deep} estimated depth using an ordinal regression approach.

Depth is generally obtained using sensors like LIDAR, Kinect, or by using stereo cameras. Sensors are expensive and stereo cameras are not generally used in endoscopy due to several restrictions. Obtaining ground-truth training data for depth estimation is very difficult in endoscopic imaging, and so, supervised methods are not feasible for endoscopic image classification. Finding correspondence between two images for 3D reconstruction is also difficult in endoscopy videos. It is quite difficult to find corresponding features across the frames.

Hence, unsupervised and semi-supervised methods are employed for MDE. Garg et al. \cite{garg2016unsupervised} used binocular stereo image pairs for the training of CNNs and then minimized a loss function formed by the wrapping of the left view image into its right of the stereo pair. Godard et al. \cite{godard2017unsupervised} improved this method by using the left-right consistency criterion. 
They trained CNNs on stereo images but used a single image for inference. They introduced a new CNN architecture that computes end-to-end MDE. The network was trained with an efficient reconstruction loss function.
The state-of-the-art unsupervised MDE method, i.e., Monodepth \cite{godard2017unsupervised} model has limited application in in-vivo images like endoscopic images. This is due to the fact that most models leverage outdoor scenes \cite{geigerwe} and a few indoor scenes \cite{saxena2007learning} for training, and they use high-end sensors or stereo cameras, while the WCE method only captures monocular images. Hence, it is important to devise a method that can perform MDE in medical imaging datasets which generally do not have ground truth depth information. That is why, a transfer learning approach is adopted in our method for estimating depth. Transfer learning refers to a learning method where what has been learned in one setting is exploited to improve generalization in another setting \cite{goodfellow2016deep}. Zero-shot learning is the extreme case of transfer learning where no labeled examples are present. In our method, a zero-shot learning approach for MDE \cite{lasinger2019towards} is employed.

The proposed method consists of two main steps. The first step focuses on depth estimation, and the second step extracts key-frames. As mentioned above, a zero-shot learning approach is adopted for depth estimation in endoscopic videos. We propose a framework to select the most informative frames of an endoscopic video sequence. Our method employs a three criteria approach to identify the key-frames. Subsequently, these key-frames can be used for 3D reconstruction. Our method is unique in a sense that it considers depth information to find key-frames. Finally, any of the selected key-frames can then be used for 3D reconstruction using a GUI. Experimental results clearly demonstrate the effectiveness of our method in selecting the key-frames, and subsequent polyp visualization.

\section{Proposed method}
\subsection{Depth estimation}
Due to insufficient ground truth depth data in endoscopy video datasets, a transfer learning approach is adopted for MDE in our proposed method. Lasinger et al. \cite{lasinger2019towards} proposed a zero-shot learning for depth estimation. 
We used a pre-trained model trained on diverse datasets by Lasinger et al. \cite{lasinger2019towards} in our work. The model was trained for depth maps obtained in three different ways. First, the dataset contains depth maps obtained using LIDAR sensors. This method gives depth maps of high quality. Second, the Structure from Motion (SfM) approach is employed to estimate the depth. The third method of getting depth information from stereo images of 3D movies dataset. It uses optical flow to find motion vectors from each of the stereo images. Then, the left-right image disparity is used to find a depth map. 

%

\begin{figure}
\centering     
\includegraphics[width=0.97\linewidth]{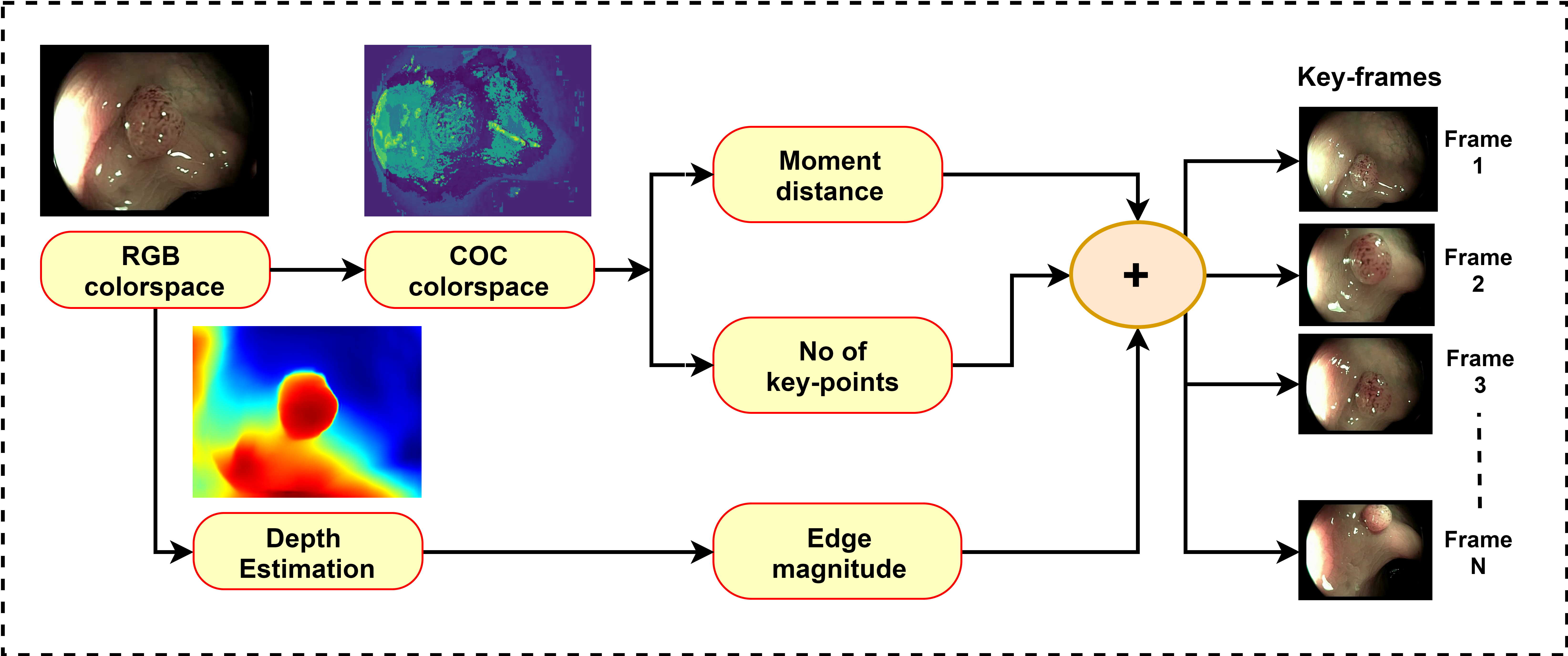}
\caption{Proposed method of finding key-frames}
\label{process}
\end{figure}


\textbf{Pre-trained network architecture}. A ResNet- based architecture as proposed by Xian et al. \cite{xian2018monocular} is used for depth estimation. Adam optimizer is used with a learning rate of    
$ 10^{-4} $ for layers which are randomly initiated and $ 10^{-5} $ for layers initialized with pre-trained weights. Decay rates for the optimizer are set at $\beta_{1}=.9$ and $ \beta_{2}=.999 $, training uses a batch size of 8. Due to different image aspect ratios, images are cropped and augmented for training.

\textbf{Loss function}. A shift and scale invariant loss function is chosen to address the problems pertaining to training on three different datasets. Let $\mathbf{d}$ $\in$  $\mathbb{R}^\mathit{N}$ be the computed inverse depth and $\mathbf{d'}$ $ \in $ $\mathbb{R}^\mathit{N}$ be ground truth inverse depth, where $\mathit{N}$ is the number of pixels in a frame. Here $s$ and $t$ represent scale and shift, respectively and they are positive real numbers. This can be represented in a vector form by taking $\vec{\mathbf{d}}_{i}$=$(\mathbf{d}_{i},1)^\intercal$ and $\mathbf{p}$=$(s,t)^\intercal$ and thus the loss function becomes: 


\begin{equation}\label{eq:a}
\begin{aligned}
\mathcal{L}(\mathbf{d}_{i},\mathbf{d'}_{i}) = \underset{\mathbf{p}}{\text{min}} && \dfrac{1}{2\mathit{N}} \sum_{i=1}^{\mathit{N}}({\vec{\mathbf{d}}_{i}}^\intercal\mathbf{p}-\mathbf{d'}_{i})^2
\end{aligned}
\end{equation}

The closed-form solution is given as:

\begin{equation}
\begin{aligned}
 \mathbf{p}^{opt}= (\sum_{i=1}^{\mathit{N}}\vec{\mathbf{d}}_{i}{\vec{\mathbf{d}}_{i}}^\intercal)^{-1}(\sum_{i=1}^{\mathit{N}}\vec{\mathbf{d}}_{i}\mathbf{d'}_{i})
\end{aligned}
\end{equation}

Substituting  $\mathbf{p}^{opt}$ into ($\ref{eq:a}$) we get:

\begin{equation}\label{eq:2}
\begin{aligned}
\mathcal{L} (\mathbf{d}_{i},\mathbf{d'}_{i}) = \underset{\mathbf{p}}{\text{min}} && \dfrac{1}{2\mathit{N}} \sum_{i=1}^{\mathit{N}}({\vec{\mathbf{d}}_{i}}^\intercal\mathbf{p}^{opt}-\mathbf{d'}_{i})^2
\end{aligned}
\end{equation}

\textbf{Regularization term}. A multi-scale scale-invariant regularization term is used which does gradient matching to the depth inverse space. This biases discontinuities to be sharp and coincide with ground truth discontinuities. 
The regularization term can be defined as,

\begin{equation}
\mathcal{L}_{r} (\mathbf{d}_{i},\mathbf{d'}_{i}) = \dfrac{1}{\mathit{N}}\sum_{j=1}^{k}\sum_{i=1}^{\mathit{N}}(|\Delta_{x} {Q_{i}}^k|+|\Delta_{y} {Q_{i}}^k|)
\end{equation}

where,
\begin{equation}
Q_{i}={\vec{\mathbf{d}}_{i}}^\intercal\mathbf{p}^{opt}-\mathbf{d'}_{i}
\end{equation}

Here $Q^{k}$ gives the difference of inverse depth maps at a scale $k$. Also, the scale is applied before finding $x$ and $y$ gradients.

\textbf{Modified loss function}. The final loss function for a training set of size $M$, taking into consideration of the regularization term becomes:

\begin{equation}
\mathcal{L}_{final} = \dfrac{1}{M}\sum_{i=1}^{M}\mathcal{L} (\mathbf{d}^{i},(\mathbf{d'})^{i})+ \alpha \mathcal{L}_{r} (\mathbf{d}^{i},(\mathbf{d'})^{i})
\end{equation}
Here $ \alpha  $ is taken as 0.5.

\subsection{Selection of key-frames}
During the colonoscopy, not all the captured frames are clinically significant. Most of the frames may have redundant information, or may not be useful from a diagnostic perspective. Such frames need to be discarded and the clinically informative frames need to be retained. 
It is also strenuous and computationally intensive for a physician to investigate each frame of a video sequence. 
Thus, we propose a key-frame selection technique. Subsequently, 3D reconstruction is done to perform further analysis of the polyps. The key-frame selection method is given in Fig.~\ref{process}.

\textbf{Colour space conversion}. Our dataset contains images which are in RGB color space. Taking clues from the human visual system which works on saliency, we changed the color space from RGB to COC which gives a better perception in the medical imaging \cite{engel1997colour}.
%
%

The image is subsequently used to find key-frames. A frame should satisfy three criteria before being selected as a key-frame. Firstly, it should be significantly different from neighboring frames. Second, the key-frame should give significant depth information of a polyp. Third, the polyp should not be occluded in the key-frame. We ensured that the above requirements are met, and they are formulated as follows:

\textbf{Image moment}: Image moments give the information of the shape of a region along with its boundaries and texture. Hu moments \cite{hu1962visual} are considered as they are invariant to affine transformation, and moment distances of consecutive frames are used to identify the redundant frames of a video.  
The frames with a higher moment distance will be considered as a key frame. The moment distance $d$ between two images is calculated as:

\begin{equation}
d=\sum_{i=1}^{i=7}{(I_{i}-I'_{i})^2}
\end{equation}

\begin{figure}[t]
\centering     
\includegraphics[width=.95\linewidth]{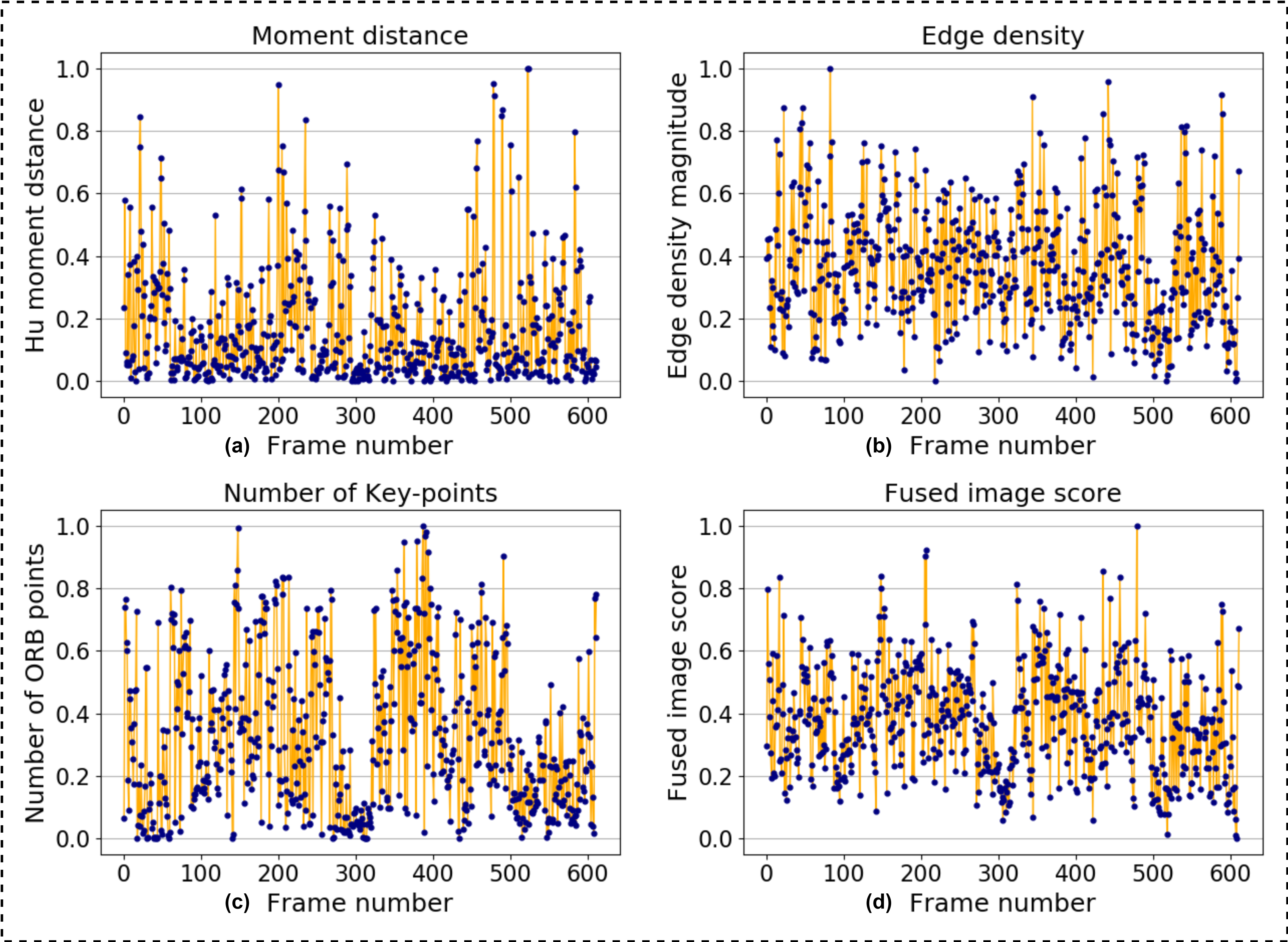}
\caption{Plot of Moment distance, Edge density, Number of key-points and the total fused score vs frame number.}
\label{Fig:chart}
\end{figure}

\textbf{Edge density}: In our proposed method, the key-frames which have significant depth information are only considered for 3D reconstruction of a polyp. It is observed that the polyp images having more edges have more depth information. The edge information can be obtained with the help of the gradient magnitude of an image. Before finding the gradients, images were smoothed using a Gaussian kernel.

Horizontal and vertical gradients are obtained using Sobel operators $S_x$ and $S_y$ and then the gradient magnitude $\Delta S$ is calculated as follows:

\begin{equation}
\Delta S=\sqrt{(S_{x})^2+(S_{y})^2}
\end{equation}

\textbf{Key-point detection}:
The proposed moment-based key-frame detection method may capture some occluded frames. So, the objective is to select non-occluded key-frames from a group of key-frames which were extracted by our proposed image moment and edge density-based criteria.  For this, a key-point detection based technique is used.

For key-point detection and extraction, we used ORB (Oriented FAST and Rotated BRIEF). ORB operates on an image to obtain a scale pyramid. It also takes into account the orientation of the image patch around the key point. Moreover, ORB is computationally faster and robust to noises in endoscopic images. The frames containing a lesser number of ORB points correspond to occluded polyps. 

\textbf{Adaptive key-frame selection}.
After finding the moment distance (d), edge magnitude (s), and the number of ORB points (p), we normalize these scores using min-max normalization.

The variable having greater variance is given more weight-age. Here, $w_i$ is the weight of the normalized score. To consider intra-variable changes, we used the sum of the magnitude of difference between consecutive frame scores as a measure to find weights. We then normalized this score to be used as weights for finding a fused score. The weights are given by:
 \begin{gather}
d_{1}=\sum_{i=1}^{n}{|d_{i}-d'_{i}|} ,
s_{1}=\sum_{i=1}^{n}{|s_{i}-s'_{i}|} ,
p_{1}=\sum_{i=1}^{n}{|p_{i}-p'_{i}|}
\end{gather}
\begin{gather}
w_{1}=\dfrac{d_{1}}{d_{1}+s_{1}+p_{1}},
w_{2}=\dfrac{s_{1}}{d_{1}+s_{1}+p_{1}}, 
w_{3}=\dfrac{p_{1}}{d_{1}+s_{1}+p_{1}}
\end{gather}
\begin{equation}
f=w_{1}d_{1}+w_{2}s_{1}+w_{3}p_{1}
\end{equation}


Here, $d_1$, $s_1$, $p_1$ are the sum of magnitudes of difference between consecutive frame scores and $f$ is the fused score obtained by adaptively weighting the three frame scores. The frames with the highest fused scores are selected according to a threshold value. The variance of each criterion with frame number is shown in Fig.~\ref{Fig:chart}.

\section{Experimental Results}
The proposed method is evaluated on the publicly available dataset. This dataset contains colonoscopic video sequences from three classes, namely adenoma, serrated and hyperplasic. The adenoma class contains 40 sequences, serrated contains 15, while hyperplasic contains 21 sequences \cite{mesejo2016computer}. In this work, we consider only the frames from the adenoma (malignant) class because this class needs the maximum attention of the physician.

\begin{figure}[t]
\centering  
\subfigure{\label{fig:1}\includegraphics[height=0.8cm,width=.1533\linewidth]{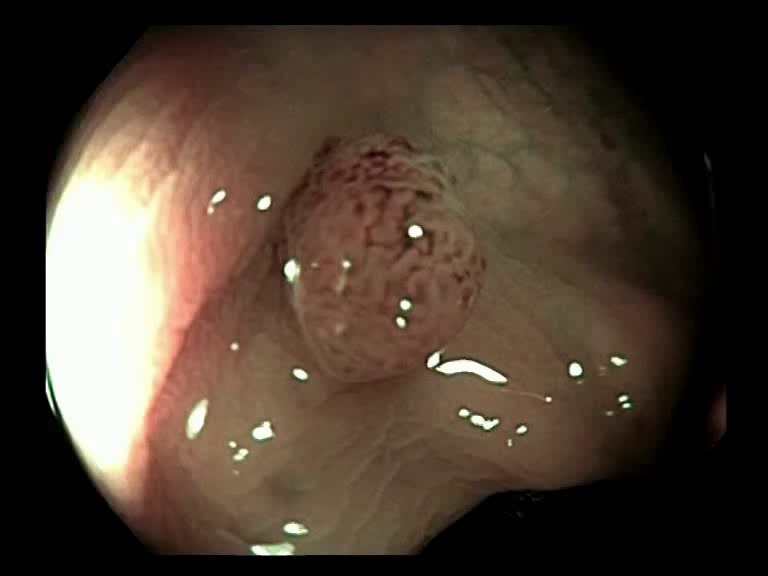}}
\subfigure{\label{fig:6}\includegraphics[height=0.8cm,width=.1533\linewidth]{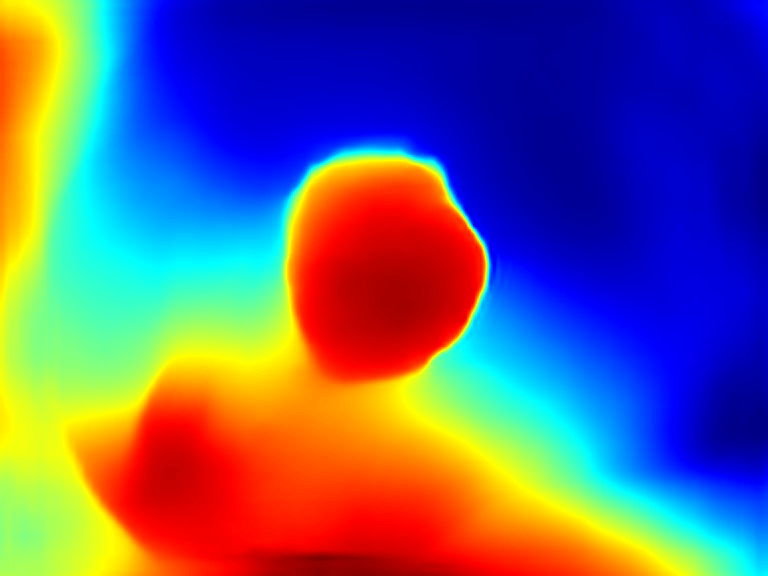}}
\subfigure{\label{fig:111}\includegraphics[height=0.8cm,width=.1533\linewidth]{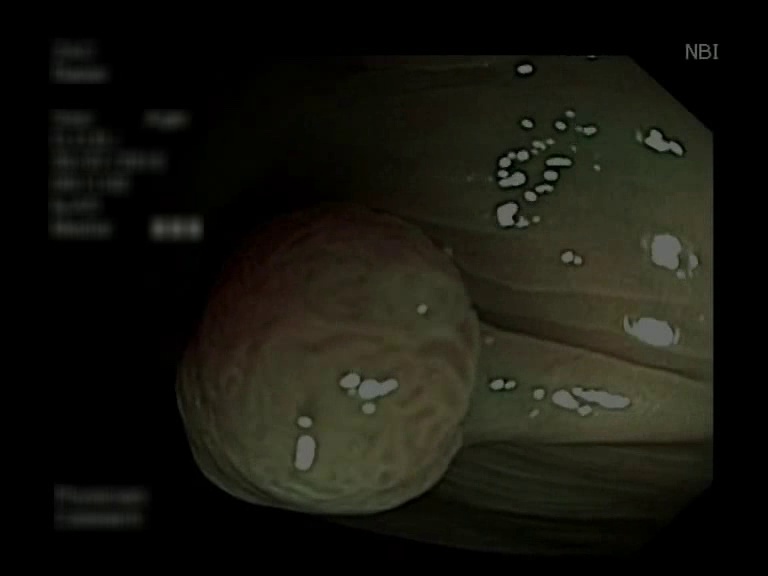}}
\subfigure{\label{fig:116}\includegraphics[height=0.8cm,width=.1533\linewidth]{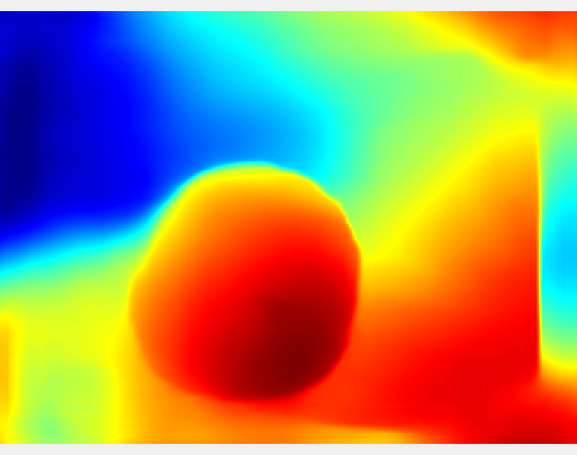}}
\subfigure{\label{fig:101}\includegraphics[height=0.8cm,width=.1533\linewidth]{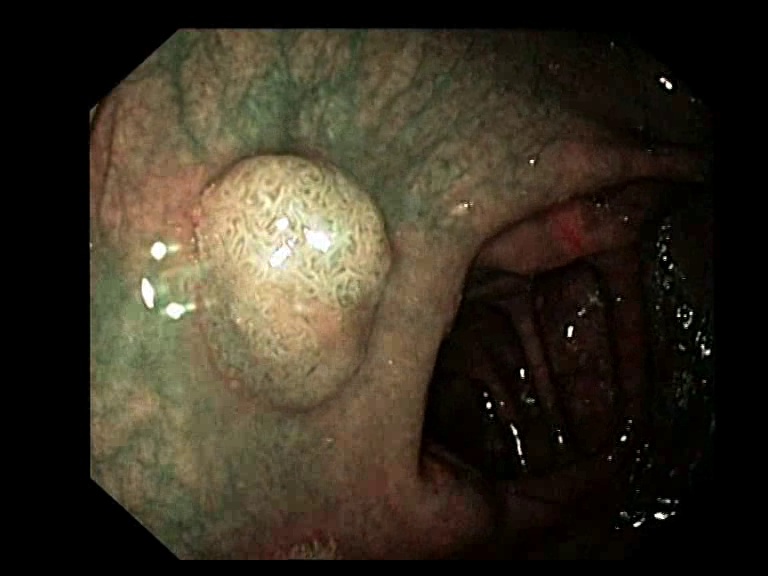}}
\subfigure{\label{fig:121}\includegraphics[height=0.80cm,width=.1533\linewidth]{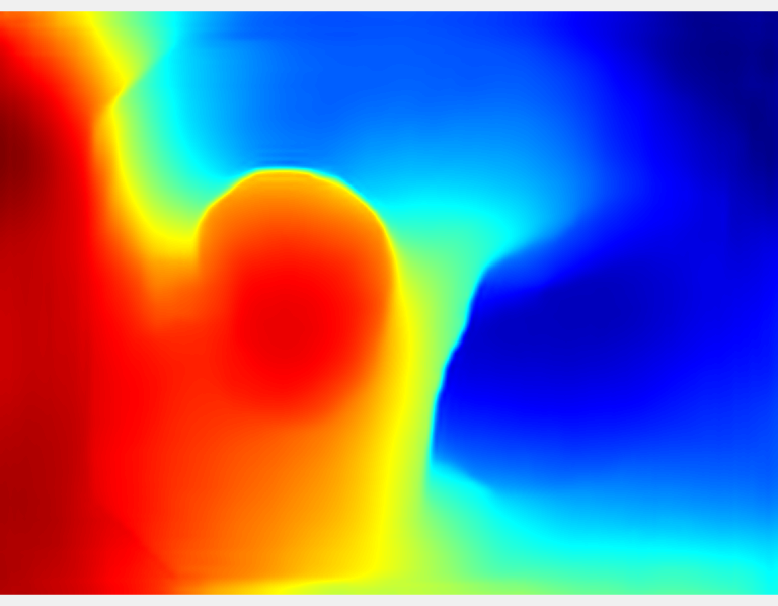}}\\

\subfigure{\label{fig:2}\includegraphics[height=0.8cm,width=.1533\linewidth]{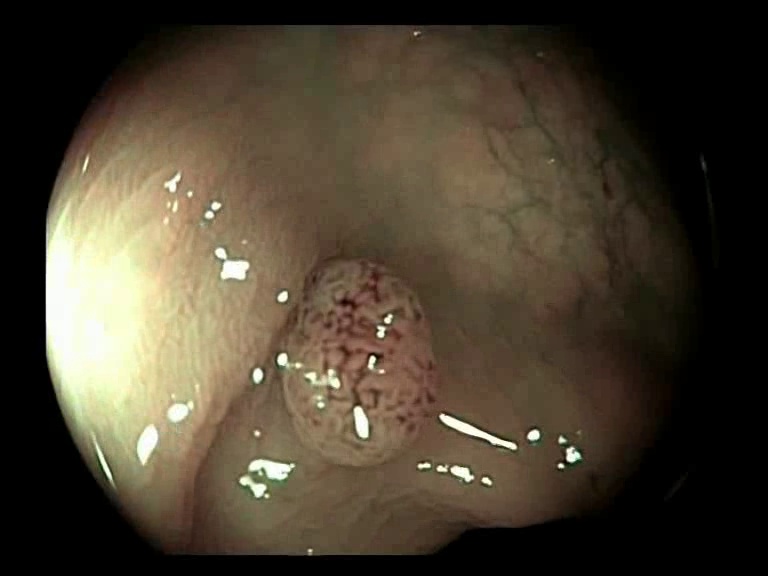}}
\subfigure{\label{fig:7}\includegraphics[height=0.8cm,width=.1533\linewidth]{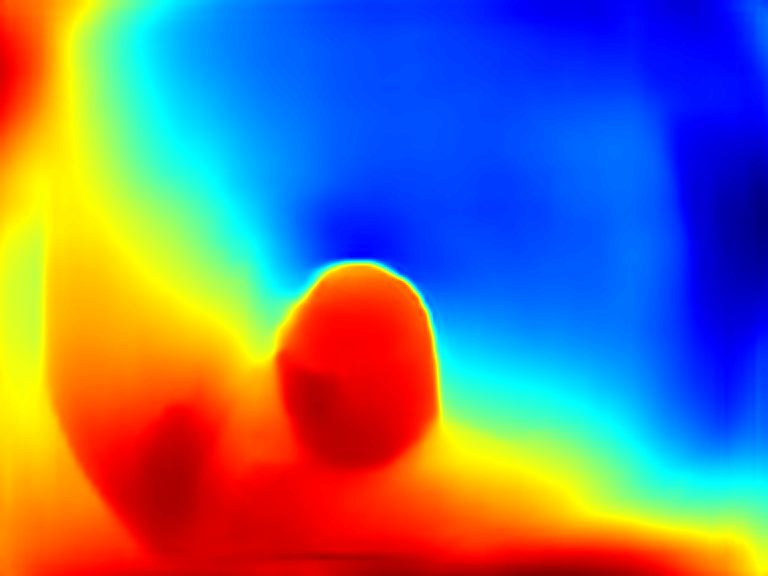}}
\subfigure{\label{fig:112}\includegraphics[height=0.8cm,width=.1533\linewidth]{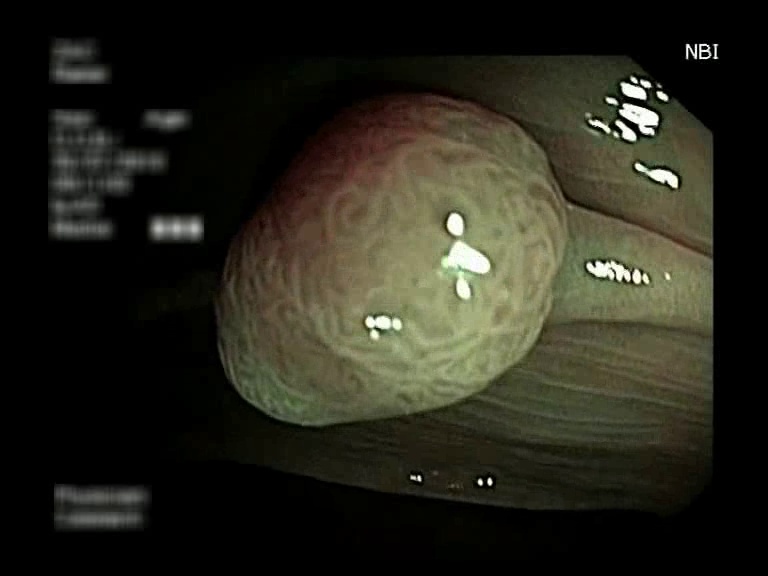}}
\subfigure{\label{fig:117}\includegraphics[height=0.8cm,width=.1533\linewidth]{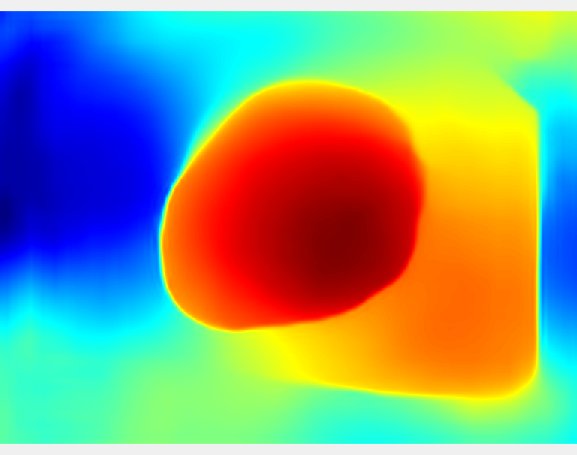}}
\subfigure{\label{fig:102}\includegraphics[height=0.8cm,width=.1533\linewidth]{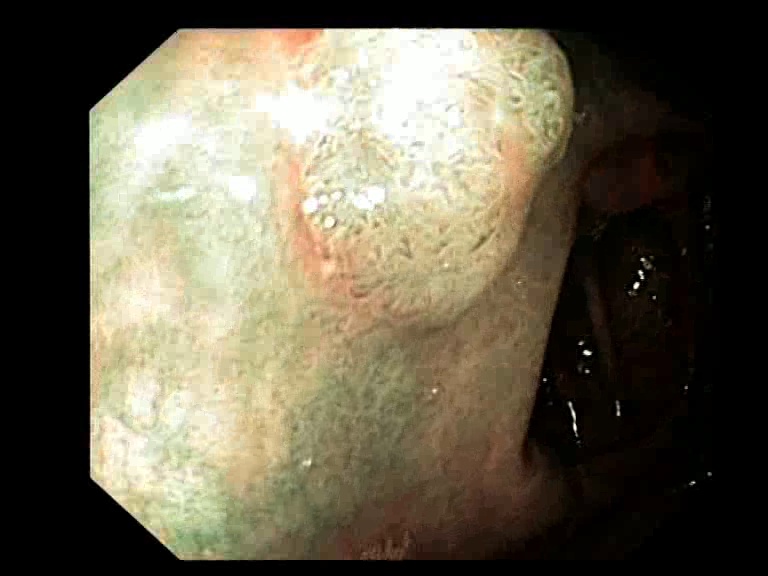}}
\subfigure{\label{fig:122}\includegraphics[height=0.8cm,width=.1533\linewidth]{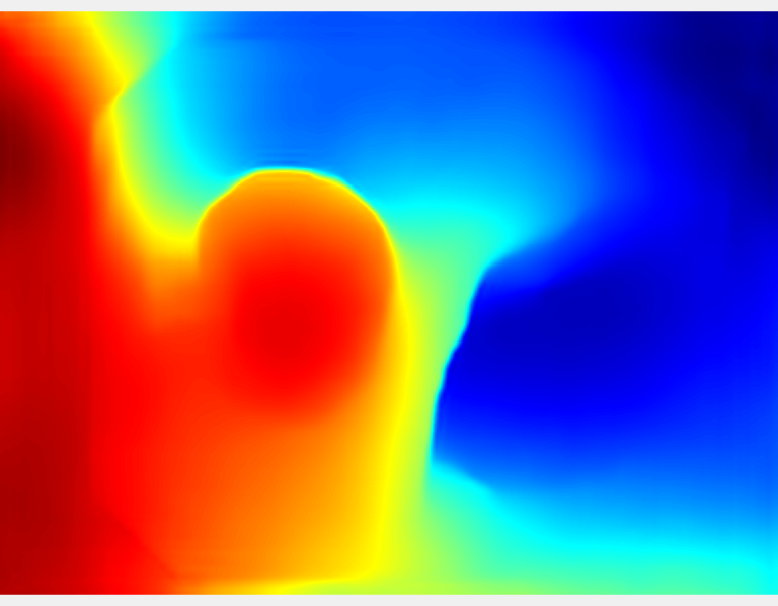}}\\

\subfigure{\label{fig:3}\includegraphics[height=0.8cm,width=.1533\linewidth]{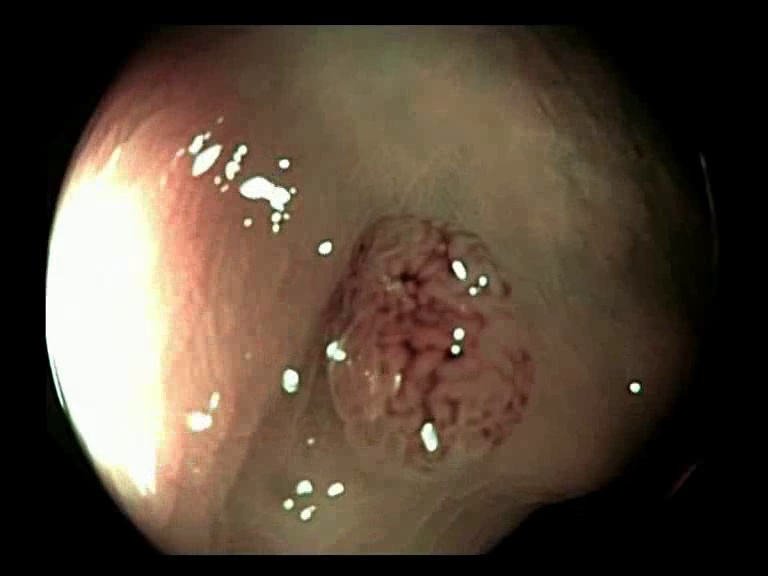}}
\subfigure{\label{fig:8}\includegraphics[height=0.8cm,width=.1533\linewidth]{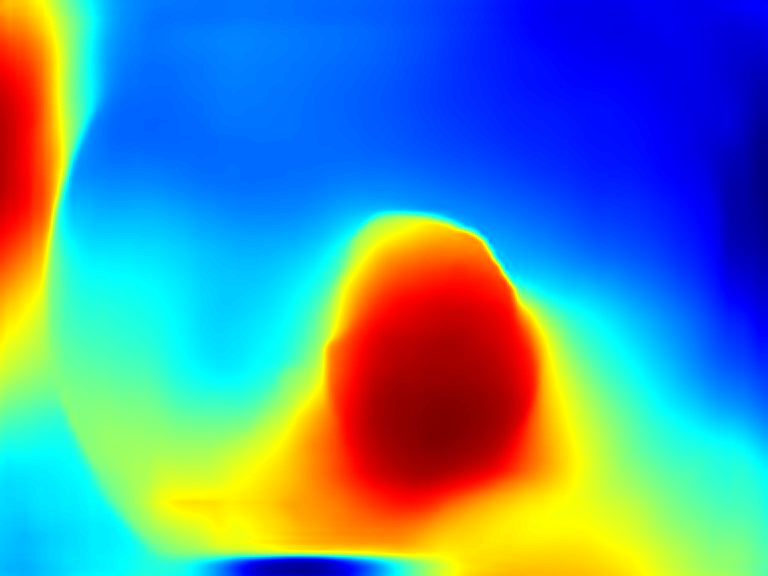}}
\subfigure{\label{fig:113}\includegraphics[height=0.8cm,width=.1533\linewidth]{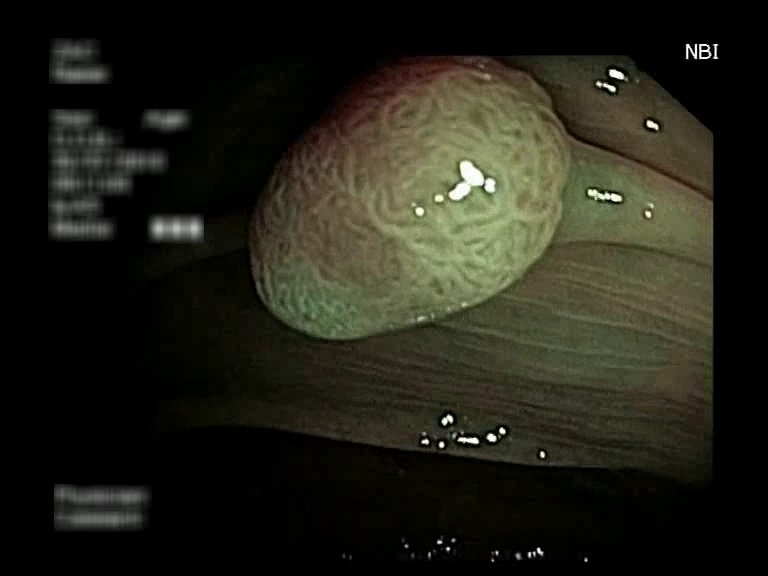}}
\subfigure{\label{fig:118}\includegraphics[height=0.8cm,width=.1533\linewidth]{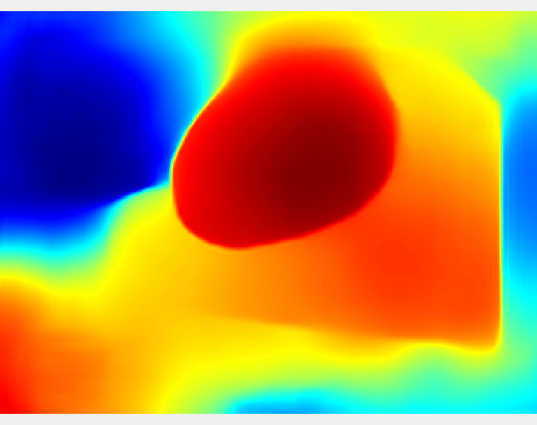}}
\subfigure{\label{fig:103}\includegraphics[height=0.8cm,width=.1533\linewidth]{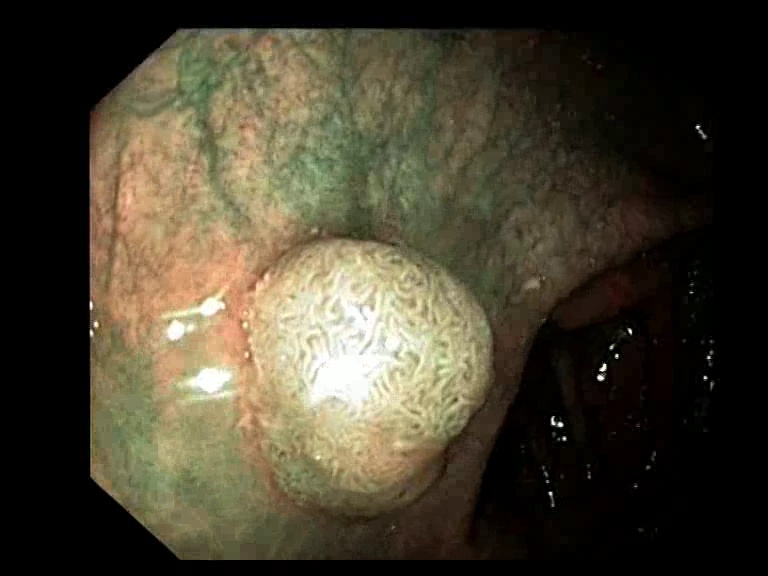}}
\subfigure{\label{fig:123}\includegraphics[height=0.8cm,width=.1533\linewidth]{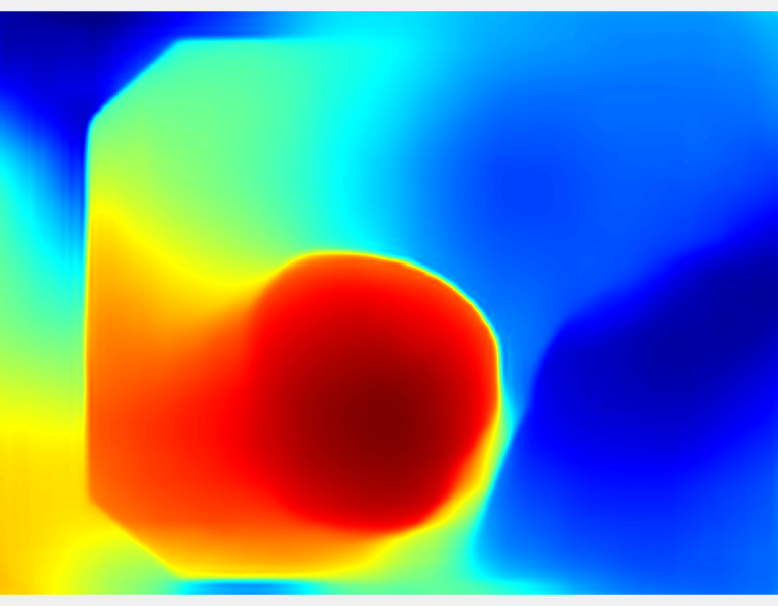}}\\

\subfigure{\label{fig:4}\includegraphics[height=0.8cm,width=.1533\linewidth]{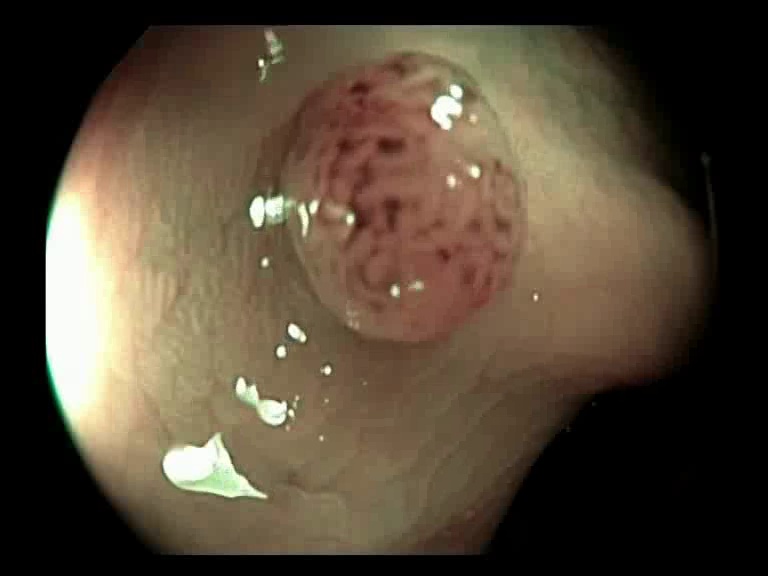}}
\subfigure{\label{fig:9}\includegraphics[height=0.8cm,width=.1533\linewidth]{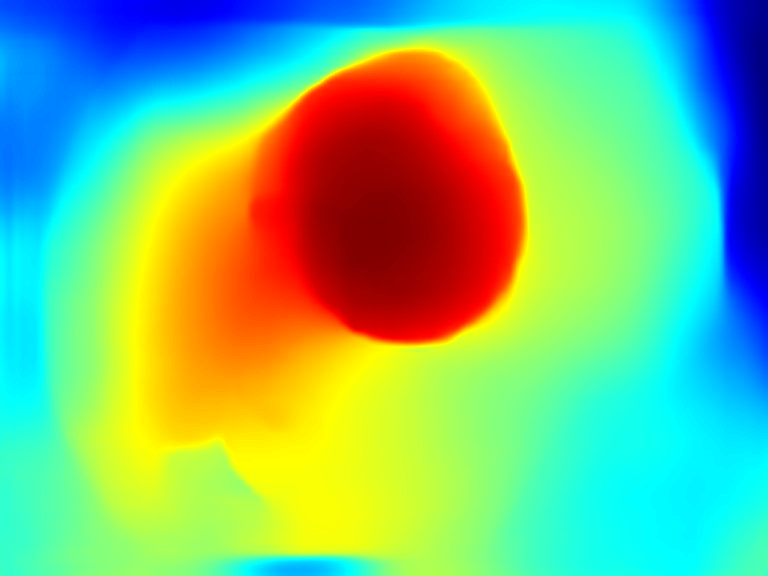}}
\subfigure{\label{fig:114}\includegraphics[height=0.8cm,width=.1533\linewidth]{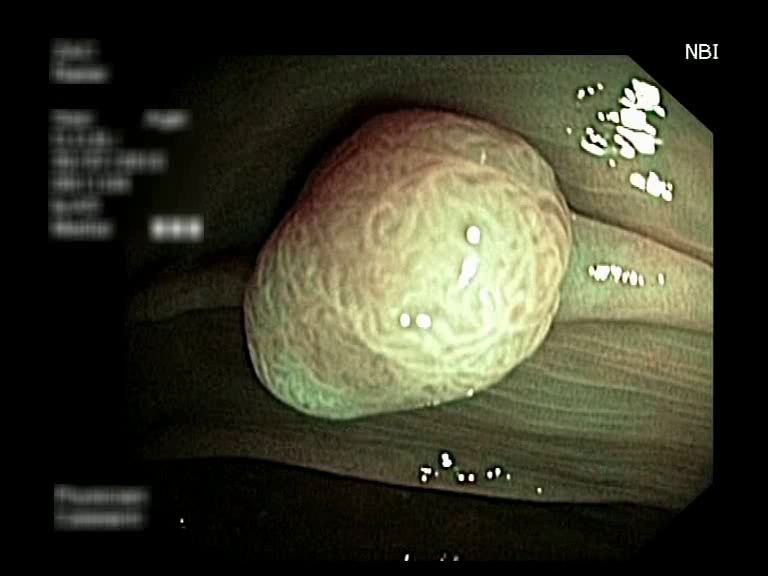}}
\subfigure{\label{fig:119}\includegraphics[height=0.8cm,width=.1533\linewidth]{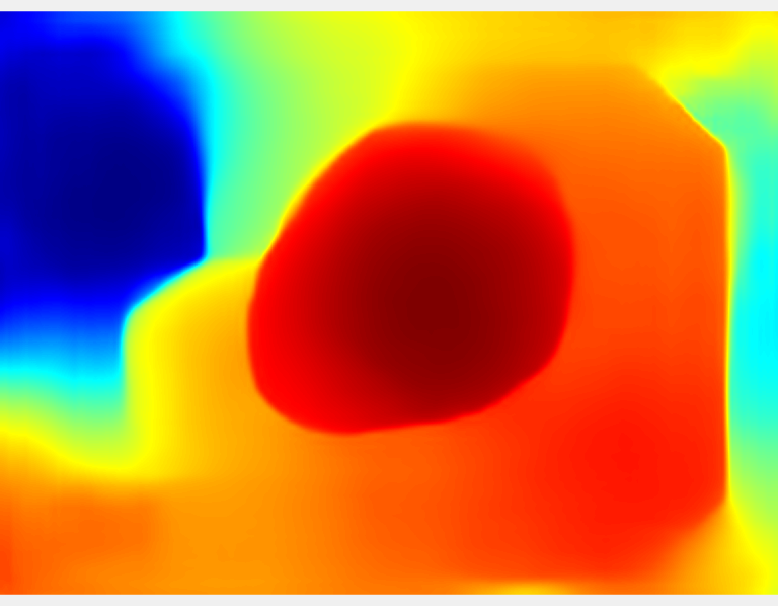}}
\subfigure{\label{fig:104}\includegraphics[height=0.8cm,width=.1533\linewidth]{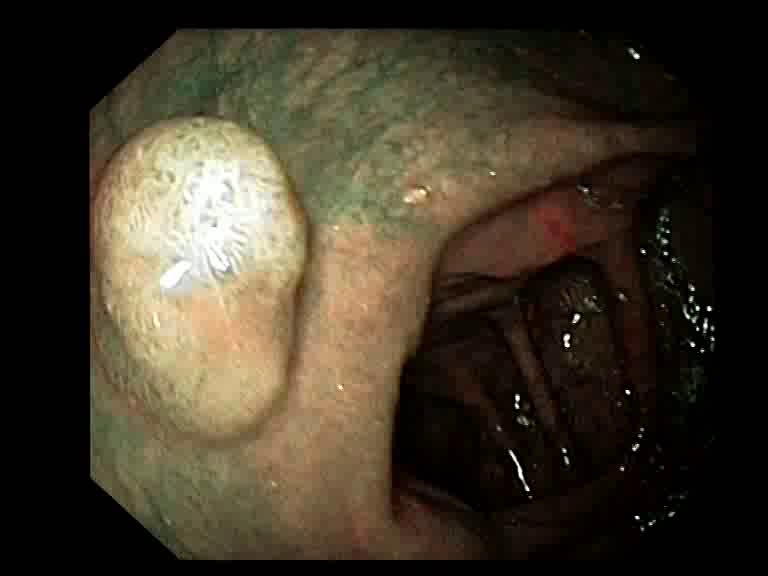}}
\subfigure{\label{fig:124}\includegraphics[height=0.8cm,width=.1533\linewidth]{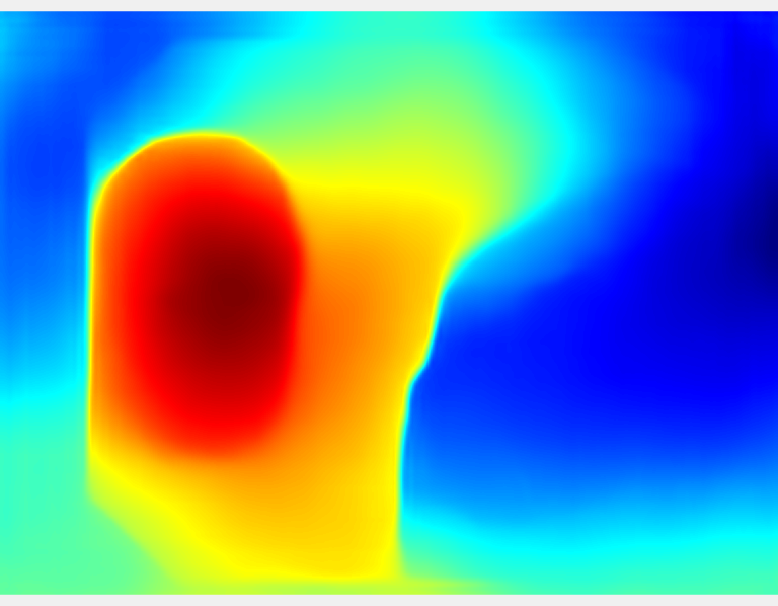}}\\

\subfigure{\label{fig:5}\includegraphics[height=0.8cm,width=.1533\linewidth]{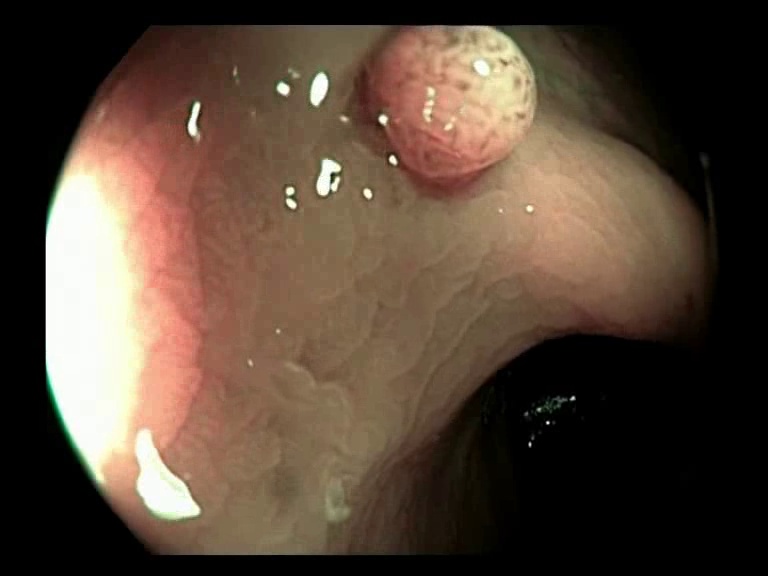}}
\subfigure{\label{fig:10}\includegraphics[height=0.8cm,width=.1533\linewidth]{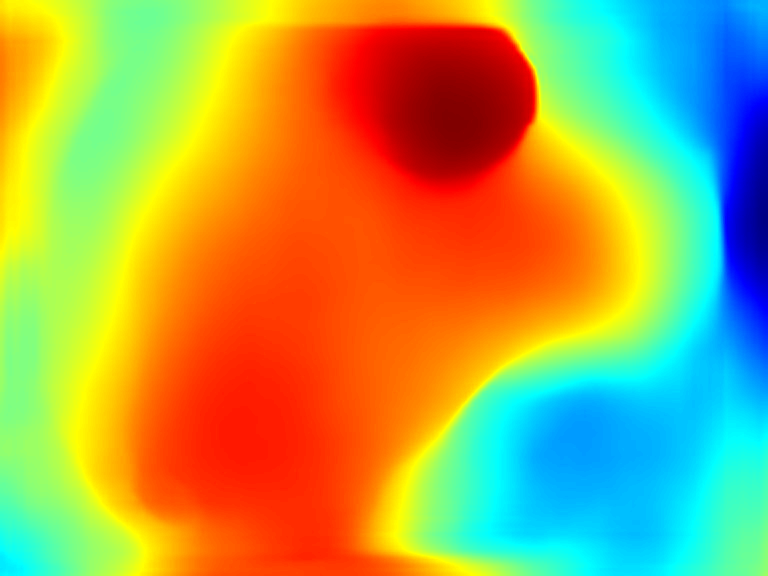}}
\subfigure{\label{fig:115}\includegraphics[height=0.8cm,width=.1533\linewidth]{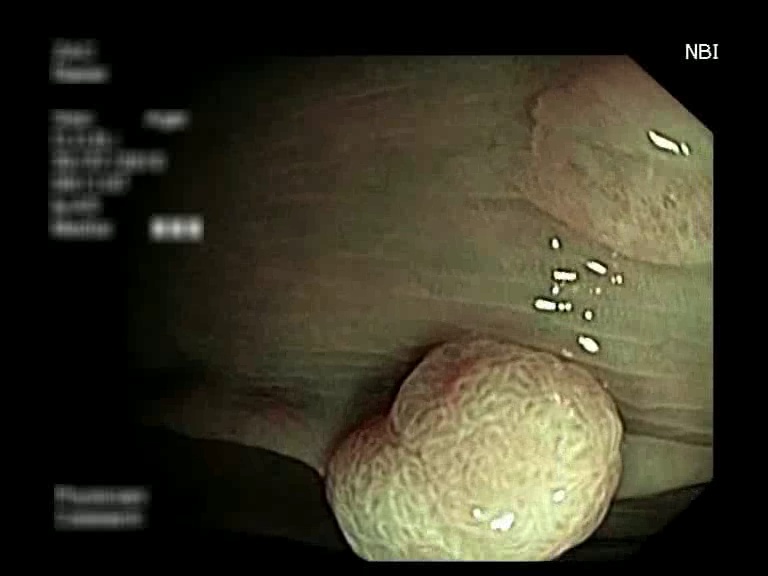}}
\subfigure{\label{fig:120}\includegraphics[height=0.8cm,width=.1533\linewidth]{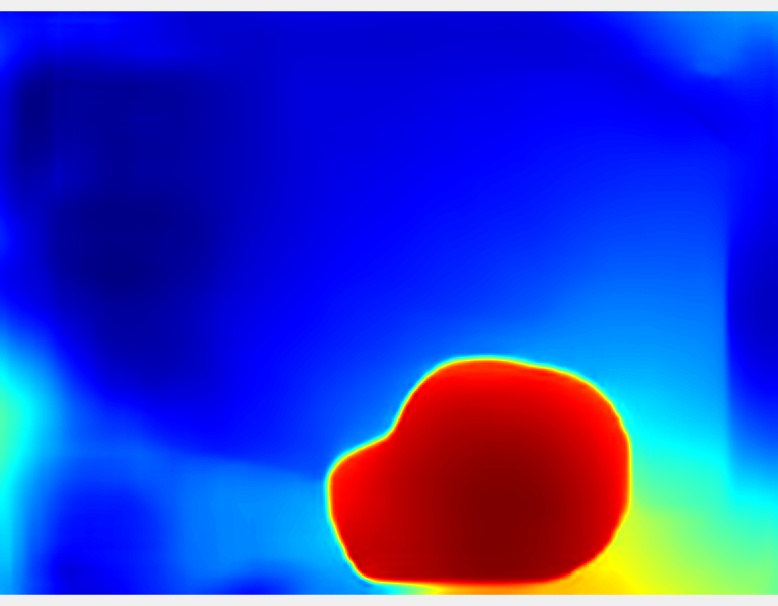}}
\subfigure{\label{fig:105}\includegraphics[height=0.8cm,width=.1533\linewidth]{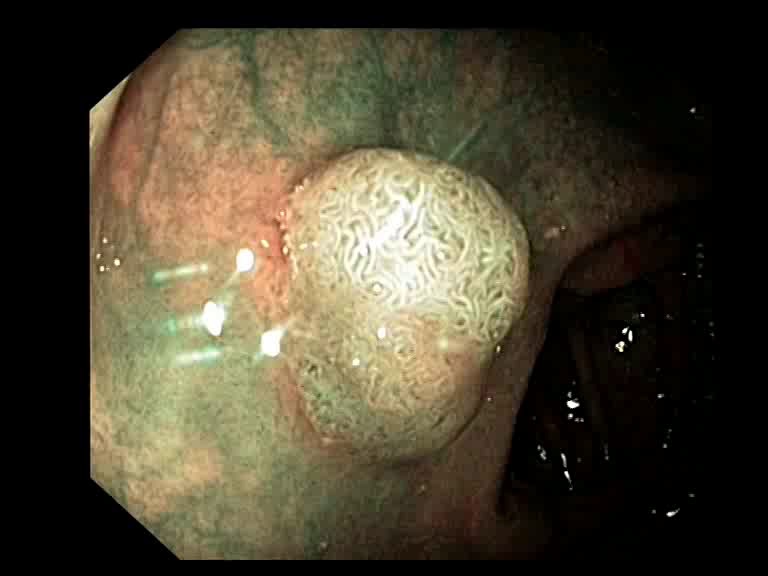}}
\subfigure{\label{fig:125}\includegraphics[height=0.8cm,width=.1533\linewidth]{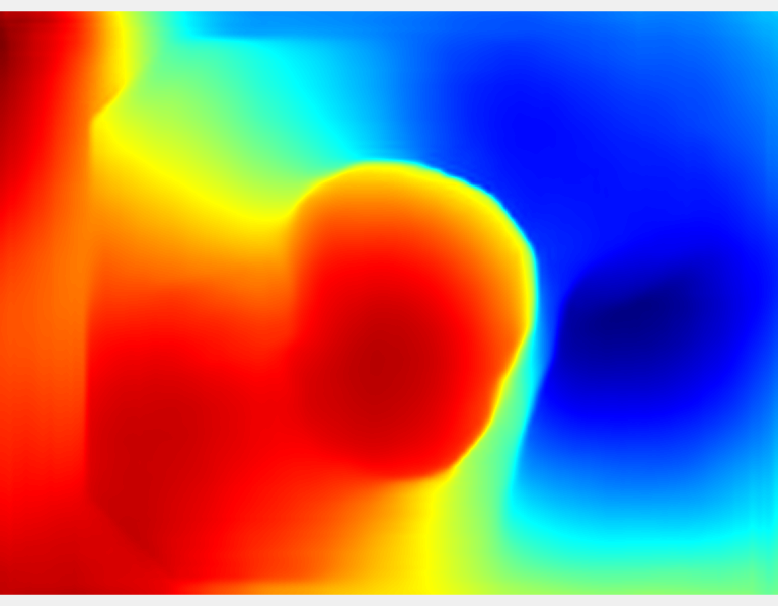}}\\

\caption{Key-frames obtained by our method and their corresponding depth maps. The polyp is visible from different viewing angles in these selected frames.}
\label{answer}
\end{figure}

\begin{figure}[h]
\centering     
\rotatebox{90}{\scriptsize Input image}
\subfigure{\label{fig:a}\includegraphics[width=.61\linewidth,height=1.4cm]{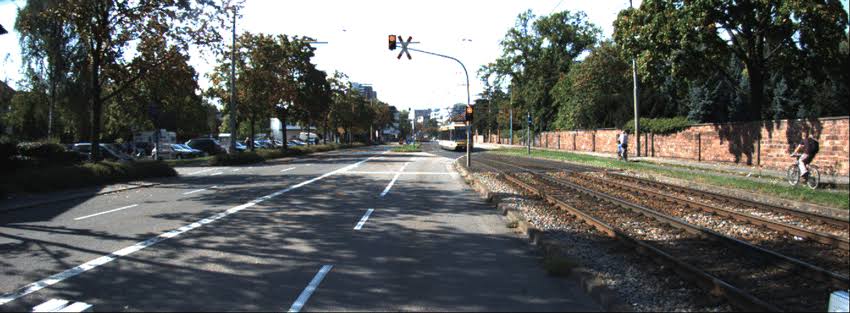}}
\subfigure{\label{fig:b}\includegraphics[width=.3\linewidth,height=1.4cm]{./IMAGES/frame36.jpg}}\\
\rotatebox{90}{\scriptsize Monodepth \cite{godard2017unsupervised}}
\subfigure{\label{fig:c}\includegraphics[width=.61\linewidth,height=1.4cm]{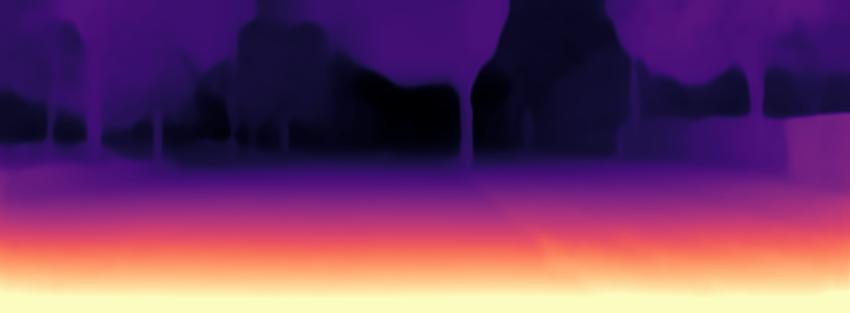}}
\subfigure{\label{fig:d}\includegraphics[width=.3\linewidth,height=1.4cm]{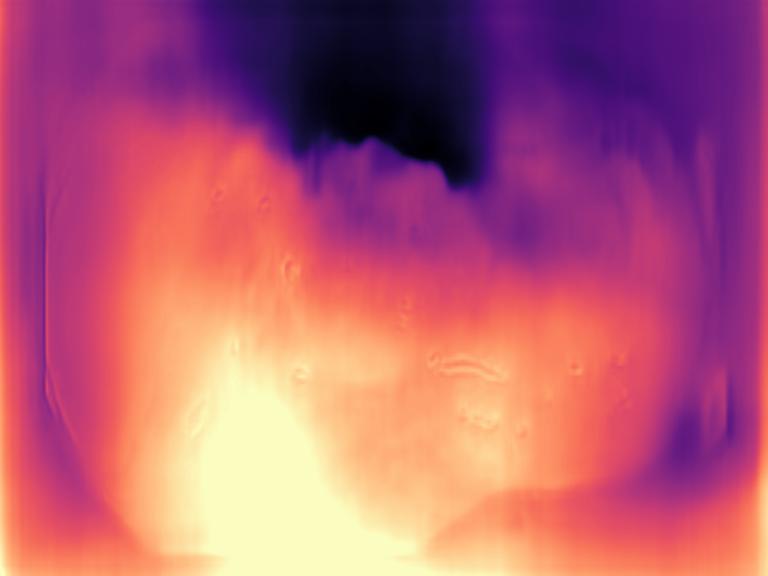}}\\
\rotatebox{90}{\scriptsize Zero-shot \cite{lasinger2019towards}}
\subfigure{\label{fig:e}\includegraphics[width=.61\linewidth,height=1.4cm]{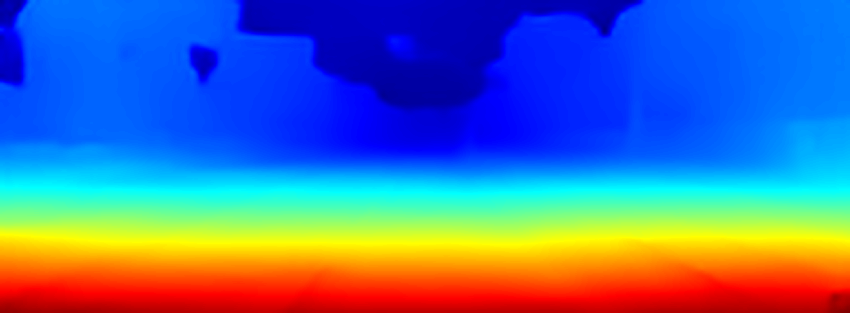}}
\subfigure{\label{fig:f}\includegraphics[width=.3\linewidth,height=1.4cm]{./IMAGES/36.png}}
\caption{Comparison of MDE on two input images, one outdoor and the other one is an endoscopy image. The depth map by Monodepth \cite{godard2017unsupervised} performs well for outdoor environment while giving unsatisfactory results for the endoscopy image . However, the zero-shot learning method \cite{lasinger2019towards} clearly performs well for medical images but cannot accurately estimate the depth in outdoor scenes. }
\label{comp}
\end{figure}

Our method performs better than the state-of-the-art MDE methods. The depth estimation results are shown in Fig.~\ref{comp}, where the first column represents the input images, while the second and the third column show the comparative results between monodepth model \cite{godard2017unsupervised} and zero-shot cross-dataset transfer pre-trained model \cite{lasinger2019towards}. This clearly shows that monodepth performs well in outdoor environments than our method. However, the Zero-shot learning method is more accurate in predicting depth in endoscopic images. 

\begin{figure}
\centering     
\includegraphics[width=\linewidth,height=5cm]{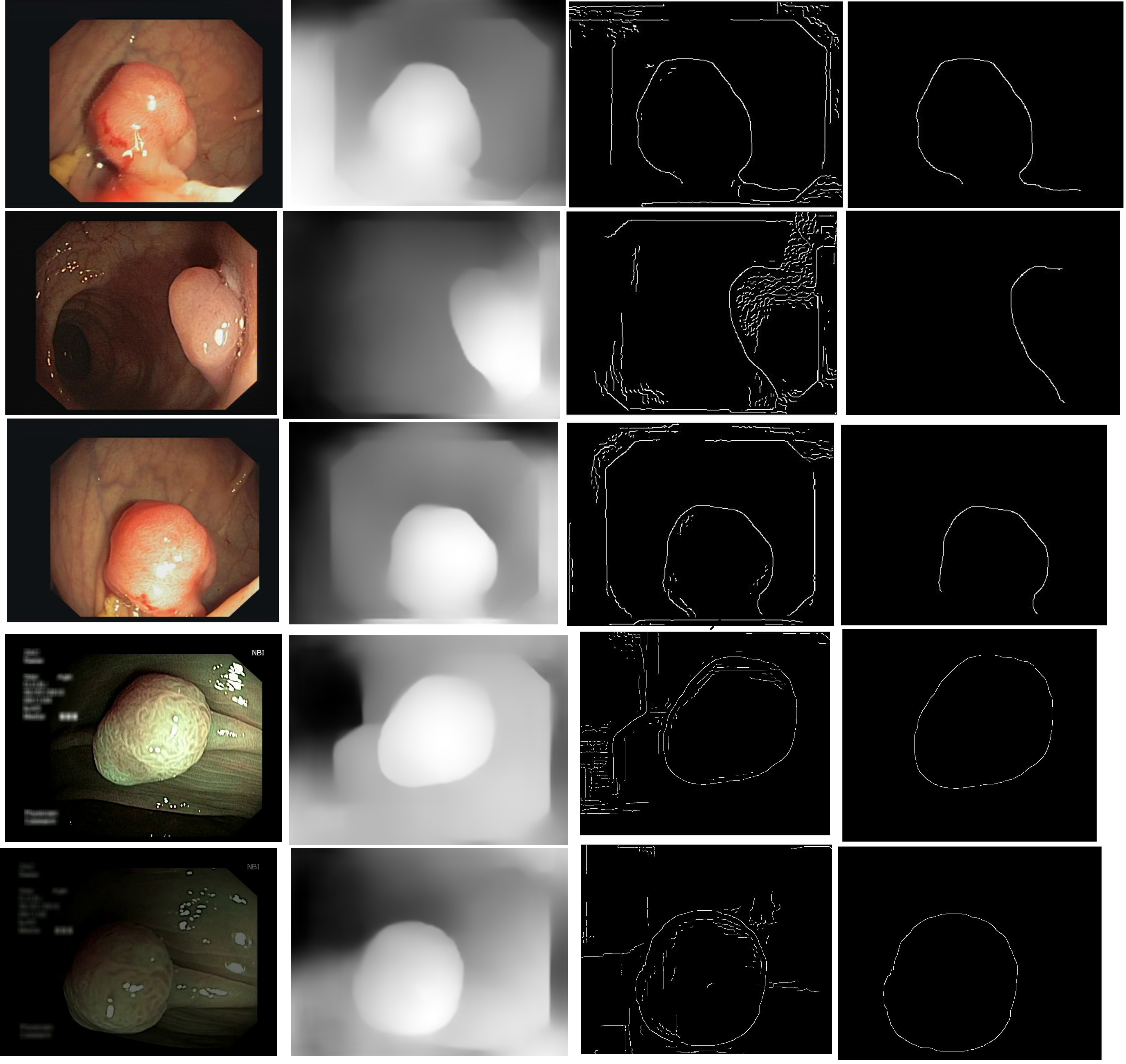}
\caption{Polyp boundary detection using depth map; Column 1: Original endoscopic image, Column 2: Generated depth maps, Column 3: Detected polyp boundary using canny edge detection algorithm, Column 4: Edge refinement using connected component analysis.  First three rows of image samples are taken from CVC clinic database \cite{bernal2015wm}, the last two rows of images are frames taken from a video sequence of the publicly available dataset \cite{mesejo2016computer}.}
\label{segmentation:examples}
\end{figure}

Our method is the first-of-its-kind in which key-frames are extracted from an endoscopic video using depth maps. Also, it is robust to occlusions. As redundant frames are discarded in our method, it is more convenient for physicians to analyze important frames of a video sequence. As explained earlier, the moment distance criterion between consecutive frames is used to ensure that redundant frames are identified, and then discarded. The edge magnitude criterion leverages the depth images data to select the best frames. Frames with fewer ORB points have occluded polyps and these frames are redundant. Adaptive thresholding is used to apply three criteria to obtain essential frames for 3D reconstruction.

The selected key-frames are finally used to reconstruct the 3D surface of the polyp. We have used Facebook's 3D image GUI to view the reconstructed polyp surface, the link to the video is shown here: 
$\href{https://youtu.be/PJKfk0Mqu2I}{https://youtu.be/PJKfk0Mqu2I}$. 3D visualization of a polyp helps in surgeries involving the removal of the polyp from its root. This gives better visualization of polyps for diagnosis. Fig.~\ref{answer} shows some of the results of key-frame extraction and the corresponding depth maps. No publicly available datasets or methods using them that predict depth maps from endoscopic frames exist. Thus, a comparison between different methods for predicting depth from endoscopic images couldn't be performed. 

Another application of our proposed method could be     
automatic segmentation of polyps in endoscopic images. The depth maps generated by our proposed method can further be used for polyp localization. The canny edge detector is used over the depth maps and subsequently, polyp boundary is determined by using connected component analysis. Fig. \ref{segmentation:examples} shows localized polyps in some of the endoscopic image samples. The segmentation performance on some of the sequences of the CVC clinic database is shown in Table \ref{tab:performance}.
 We defined mIoU as the mean intersection over the union of the segmented polyp masks to the ground truth masks. In polyp segmentation, an IoU score of $\geq$ 0.5 is generally considered good \cite{yamada2019development}. 
\begin{table}[]
\centering
\caption{Key frame selection and segmentation performance using our method on some of the sequences of CVC clinic database (Sequences with only the elevated polyps are considered)}
\begin{tabular}{ccc}
\hline\hline
\multicolumn{1}{l}{\textbf{Sequence}} & \multicolumn{1}{l}{\textbf{\#Key frames}}  & \multicolumn{1}{l}{\textbf{mIoU $>$ 0.5?}} \\ \hline
104-126                               & 7                                        & Yes                                      \\ \hline
127-151                               & 11                                       & Yes                                      \\ \hline
298-317                               & 2                                        & Yes                                      \\ \hline
343-363                               & 7                                        & Yes                                      \\ \hline
384-408                               & 13                                       & Yes                                      \\ \hline
409-428                               & 8                                        & Yes                                      \\ \hline
479-503                               & 20                                       & Yes                                      \\ \hline
504-528                               & 6                                        & Yes                                      \\ \hline
572-591                               & 4                                        & Yes                                      \\ \hline
592-612                               & 5                                        & Yes                                      \\ \hline
\end{tabular}
\label{tab:performance}
\end{table}
\section{Conclusion}

Our proposed method can determine depth maps using a zero-shot learning approach. 
The essential frames are picked out from WCE videos with the help of depth information and the proposed three criteria selection strategy. The selection of a threshold value for the final fused score must be empirically set to extract the key-frames. Experimental results show the efficacy of the proposed method in selecting key frames from endoscopic videos and subsequent segmentation of detected polyps in the key frames with the help of extracted depth maps.
Also, the 3D model could be used in clinical diagnosis and surgeries.
One possible extension of this work could be the visualization of polyps in detected key frames in an augmented reality framework.

\section*{Acknowledgement}
The work of Yuji Iwahori was supported by the JSPS Grant-in-Aid Scientific Research (C) under Grant 20K11873 for the endoscope and other medical image researches

\bibliographystyle{IEEEtran}
\balance
\bibliography{Reference_1}

%
%




\end{document}